\documentclass[11pt]{article}
\usepackage{acl2014}
\usepackage{times}
\usepackage{url}
\usepackage{latexsym}
\usepackage{graphicx}
\usepackage{color}
\usepackage{subcaption}

\newcommand{\drLEX}{\nobreak{DR-{\sc lex}$_{1}$}}
\newcommand{\dr}{\nobreak{DR-{\sc nolex}}}
\newcommand{\drLEXto}{\nobreak{DR-{\sc lex}$_{2}$}}
\newcommand{\drLEXoo}{\nobreak{DR-{\sc lex}$_{1.1}$}}

\newcommand{\drLEXtt}{\nobreak{DR-{\sc lex}$_{2.1}$}}
\newcommand{\disco}{\nobreak{DiscoTK}}
\newcommand{\discolight}{\nobreak{{\sc DiscoTK}$_{light}$}}
\newcommand{\discoparty}{\nobreak{{\sc DiscoTK}$_{party}$}}
\newcommand{\spede}{\nobreak{\sc spede07pP}}
\newcommand{\sempos}{\nobreak{SEMPOS}}

\newcommand{\asiya}{\nobreak{\sc Asiya}}

\newcommand{\Ni}{({\em i})~}
\newcommand{\Nii}{({\em ii})~}
\newcommand{\Niii}{({\em iii})~}
\newcommand{\Niv}{({\em iv})~}
\newcommand{\Nv}{({\em v})~}

\include{amssym}

\title{DiscoTK: Using Discourse Structure for Machine Translation Evaluation}

\author{Shafiq Joty\hspace*{3mm}
Francisco Guzm\'an\hspace*{3mm}
Llu\'is M\`arquez\hspace*{2mm}\hbox{\rm and}\hspace*{2mm}Preslav Nakov\\
ALT Research Group\\
Qatar Computing Research Institute --- Qatar Foundation\\
{\tt\{sjoty,fguzman,lmarquez,pnakov\}@qf.org.qa}}

\date{}

\begin{document}

\maketitle

\begin{abstract}

We present novel automatic metrics for machine translation evaluation
that use discourse structure and convolution kernels
to compare 
the discourse tree of an automatic translation with that of the human reference.
We experiment with five transformations and augmentations
of a base discourse tree representation based on the rhetorical structure theory,
and we combine the kernel scores for each of them into a single score.
Finally, we add other metrics from the \asiya\ MT evaluation toolkit,
and we tune the weights of the combination on actual human judgments.
Experiments on the WMT12 and WMT13 metrics shared task datasets
show correlation with human judgments that outperforms
what the best systems that participated in these years achieved,
both at the segment and at the system level.



\end{abstract}


\section{Introduction}
\label{sec:intro}
The rapid development of statistical machine translation (SMT) that we have seen in recent years
would not have been possible without automatic metrics for measuring SMT quality.
In particular, the development of BLEU~\cite{Papineni:Roukos:Ward:Zhu:2002}
revolutionized the SMT field,
allowing not only to compare two systems in a way that strongly correlates with human judgments,
but it also enabled the rise of discriminative log-linear models,
which use optimizers such as MERT~\cite{och03minimum},
and later MIRA~\cite{watanabe-EtAl:2007,Chiang:2008} and PRO~\cite{Hopkins2011},
to optimize BLEU, or an approximation thereof, directly.
While over the years other strong metrics such as TER~\cite{Snover06astudy} and Meteor~\cite{Lavie:2009:MMA} have emerged, BLEU remains the de-facto standard, despite its simplicity.

Recently, there has been steady increase in BLEU scores for well-resourced language pairs such as Spanish-English and Arabic-English.
However, it was also observed that BLEU-like $n$-gram matching metrics
are unreliable for high-quality translation output \cite{Doddington:2002:AEM,lavie-agarwal:2007:WMT}.
In fact, researchers already worry that BLEU will soon be unable to distinguish
automatic from human translations.\footnote{
This would not mean that computers have achieved human proficiency;
it would rather show BLEU's inadequacy.}
This is a problem for most present-day metrics,
which cannot tell apart raw machine translation output
from a fully fluent professionally post-edited version thereof~\cite{denkowski2012challenges}.

Another concern is that BLEU-like $n$-gram matching metrics
tend to favor phrase-based SMT systems over rule-based systems and other SMT paradigms. 
In particular, they are unable to capture the syntactic and semantic structure of sentences,
and are thus insensitive to improvement in these aspects.
%
Furthermore,
it has been shown that lexical similarity is both insufficient and not strictly necessary
for two sentences to convey the same meaning~\cite{Culy:2003,Coughlin:2003,Callison-Burch:2006}.

The above issues have motivated a large amount of work dedicated to design better evaluation metrics.
The Metrics task at the Workshop on Machine Translation (WMT)
has been instrumental in this quest.
Below we present QCRI's submission to the Metrics task of WMT14,
which consists of the \disco\ family of discourse-based metrics.


In particular, we 
experiment with five different transformations and augmentations of a discourse tree representation, and we combine the kernel scores for each of them into a single score which we call \discolight.
Next, we add to the combination other metrics from the \asiya\ MT evaluation toolkit \cite{AsiyaMTjournal:2010}, to produce the \discoparty\ metric.

Finally, we tune the relative weights of the metrics in the combination using human judgments
in a learning-to-rank framework.
This proved to be quite beneficial:
the tuned version of the \discoparty \ metric was the best performing metric in the WMT14 Metrics shared task.

The rest of the paper is organized as follows:
Section~\ref{sec:DRmetrics} introduces our basic discourse metrics and the tree representations they are based on.
Section~\ref{sec:combination} describes our metric combinations.
Section~\ref{sec:experiments} presents our experiments and results on datasets from previous years.
Finally, Section~\ref{sec:conclusion} concludes and suggests directions for future work.


\section{Discourse-Based Metrics}
\label{sec:DRmetrics}
\begin{figure*}[bht]
  \begin{subfigure}{.8\linewidth}
  \mbox{\hspace{0.0cm}
  \includegraphics[scale=0.7]{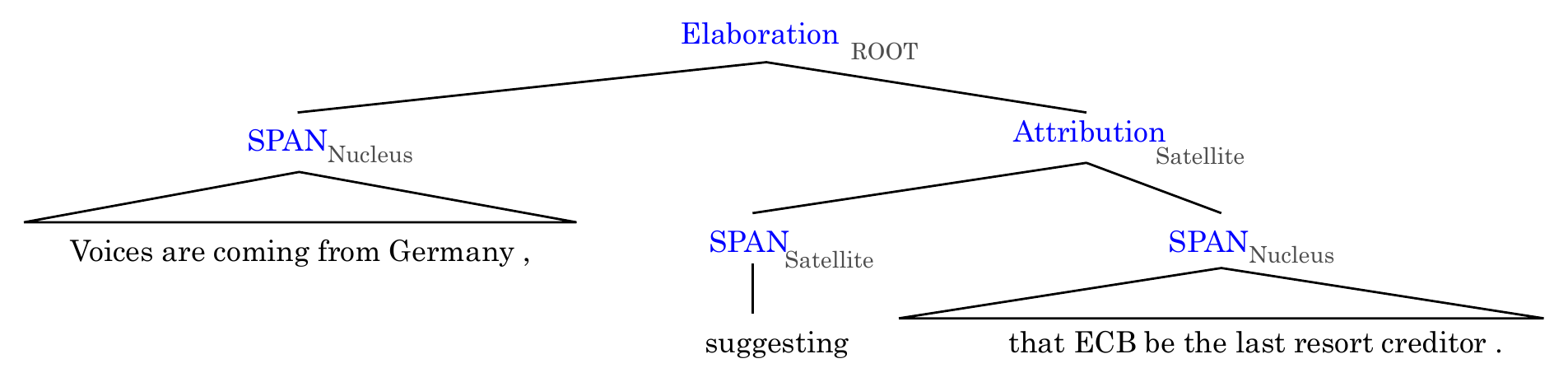}}
  \caption{A reference (human-written) translation.}
  \end{subfigure}

  \begin{subfigure}{.5\linewidth}
	\mbox{\hspace{0.0cm}
		 \includegraphics [scale=0.6] {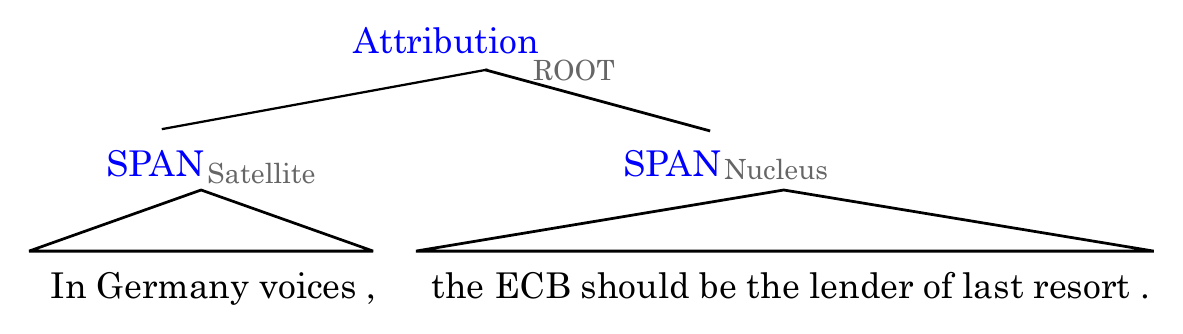}}
	    \caption{A higher quality (system-generated) translation.}
	     \label{fig:lex2}
	\end{subfigure}
	\hspace{0.3cm}
	\begin{subfigure}{.5\linewidth}
	\mbox{\hspace{-0.6cm}
		 \includegraphics [scale=0.6] {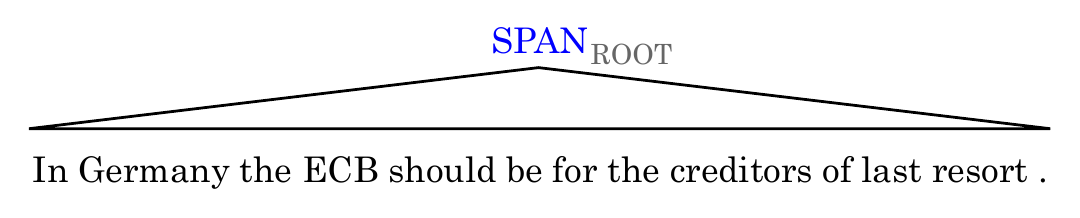}}
		  \caption{A lower quality (system-generated) translation.}
	     \label{fig:lex1.1}
	\end{subfigure}
  \caption{Three discourse trees for the translations of a source sentence:
          (a) the reference, (b) a higher quality automatic translation, and (c) a lower quality automatic translation.} 
  \label{fig:DTs}
\end{figure*}

In our recent work \cite{discoMT:acl2014}, we used the information embedded in the discourse-trees (DTs) to compare the output of an MT system to a human reference. More specifically, we used a state-of-the-art sentence-level discourse parser \cite{Shafiq12} to generate discourse trees for the sentences in accordance with the Rhetorical Structure Theory (RST) of discourse \cite{Mann88}. Then, we computed the similarity between DTs of the human references and the system translations using a convolution tree kernel \cite{Collins01}, which efficiently computes the number of common subtrees. Note that this kernel was originally designed for syntactic parsing, and the subtrees are subject to the constraint that their nodes are taken with all or none of their children, i.e., if we take a direct descendant of a given node, we must also take all siblings of that descendant. This imposes some limitations on the type of substructures that can be compared, and motivates the enriched tree representations explained in subsections 2.1--2.4.


The motivation to compare discourse trees, is that translations should preserve the coherence relations. For example, consider the three discourse trees (DTs) shown in Figure~\ref{fig:DTs}. Notice that the \emph{Attribution} relation in the reference translation is also realized in the system translation in (b) but not in (c), which makes (b) a better translation compared to (c), according to our hypothesis. 

In \cite{discoMT:acl2014}, we have shown that
discourse structure provides additional 
information for MT evaluation,
which is not captured by existing metrics that use lexical, syntactic and semantic information;
thus, discourse should be considered when developing new rich metrics.
%

Here, we extend our previous work by developing metrics that are based on new representations of the DTs.
In the remainder of this section, we will focus on the individual DT representations that we will experiment with;
then, the following section will describe the metric combinations and tuning used to produce the \disco~metrics.

\subsection{\drLEX}

Figure \ref{fig:lex1} shows our first representation of the DT. The lexical items, i.e., words, constitute the leaves of the tree. The words in an Elementary Discourse Unit (EDU) are grouped under a predefined tag \textbf{EDU}, to which the nuclearity status of the EDU is attached: \emph{nucleus} vs. \emph{satellite}. Coherence relations, such as \emph{Attribution}, \emph{Elaboration}, and \emph{Enablement}, between adjacent text spans constitute the internal nodes of the tree. Like the EDUs, the nuclearity statuses of the larger discourse units are attached to the relation labels. Notice that with this representation the tree kernel can easily be extended to find subtree matches at the word level, i.e., by including an additional layer of \emph{dummy} leaves as was done in \cite{Moschitti07}. We applied the same solution in our representations. 


\subsection{\dr}

Our second representation \dr \ (Figure \ref{fig:nolex}) is a simple variation of \drLEX , where we exclude the lexical items. This allows us to measure the similarity between two translations in terms of their discourse structures alone.

\subsection{\drLEXto}

One limitation of \drLEX \ and \dr \ is that they do not separate the structure, i.e., the skeleton, of the tree from its labels. Therefore, when measuring the similarity between two DTs, they do not allow the tree kernel to give partial credit to subtrees that differ in labels but match in their structures. \drLEXto , a variation of \drLEX , addresses this limitation as shown in Figure \ref{fig:lex2}. It uses predefined tags \textbf{SPAN} and \textbf{EDU} to build the skeleton of the tree, and considers the nuclearity and/or relation labels as properties (added as children) of these tags. For example, a \textbf{SPAN} has two properties, namely its nuclearity and its relation, and an \textbf{EDU} has one property, namely its nuclearity. The words of an EDU are placed under the predefined tag \textbf{NGRAM}.

\subsection{\drLEXoo \ and \drLEXtt}

\begin{figure*}[t]
	\begin{subfigure}{.5\linewidth}
 		\mbox{ \hspace{-0.0cm}
		\includegraphics [scale=0.50] {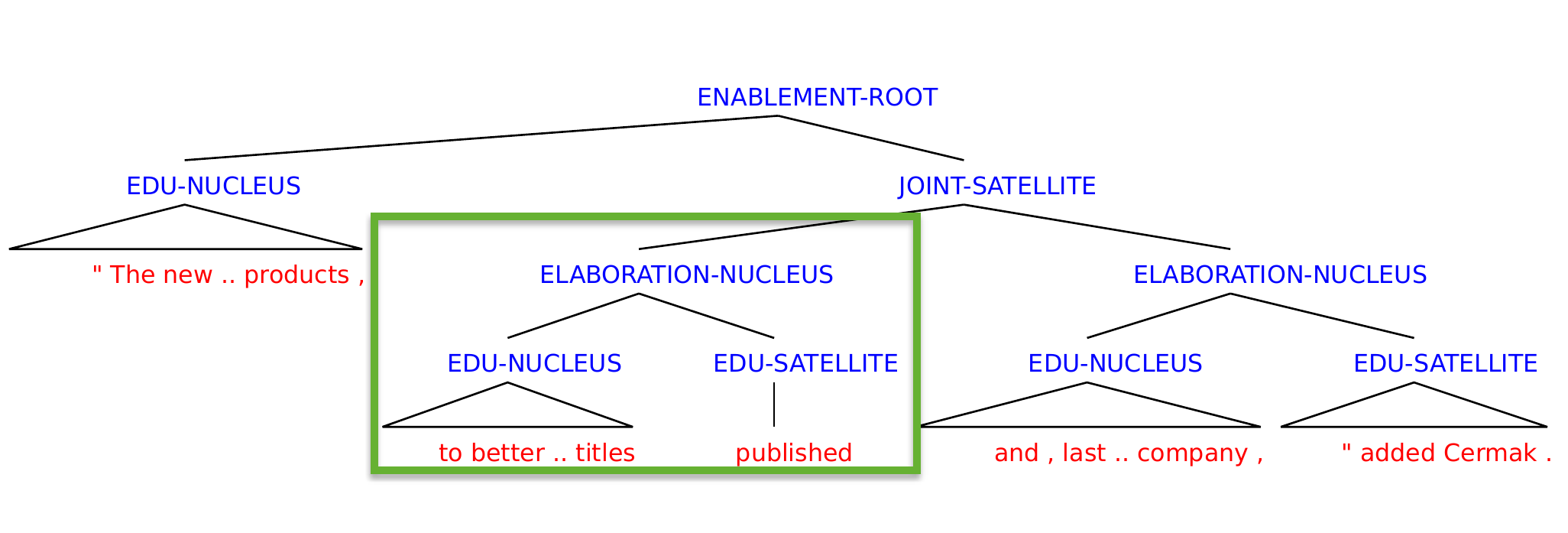}}
	    \caption{DT for \drLEX .}
	     \label{fig:lex1}
	\end{subfigure}	\hspace{1.3cm}
	\begin{subfigure}{.5\linewidth}
	\centering
		\includegraphics [scale=0.50] {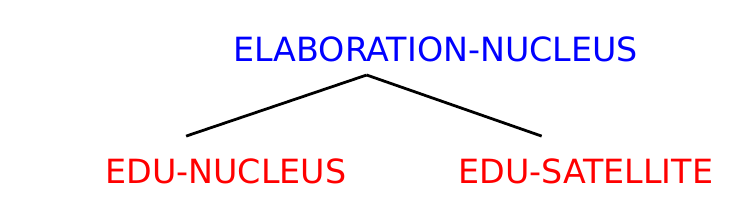}
	    \caption{DT for \dr .}
	     \label{fig:nolex}
	\end{subfigure}

	\begin{subfigure}{.5\linewidth}
	\mbox{\hspace{-0.4cm}
		 \includegraphics [scale=0.45] {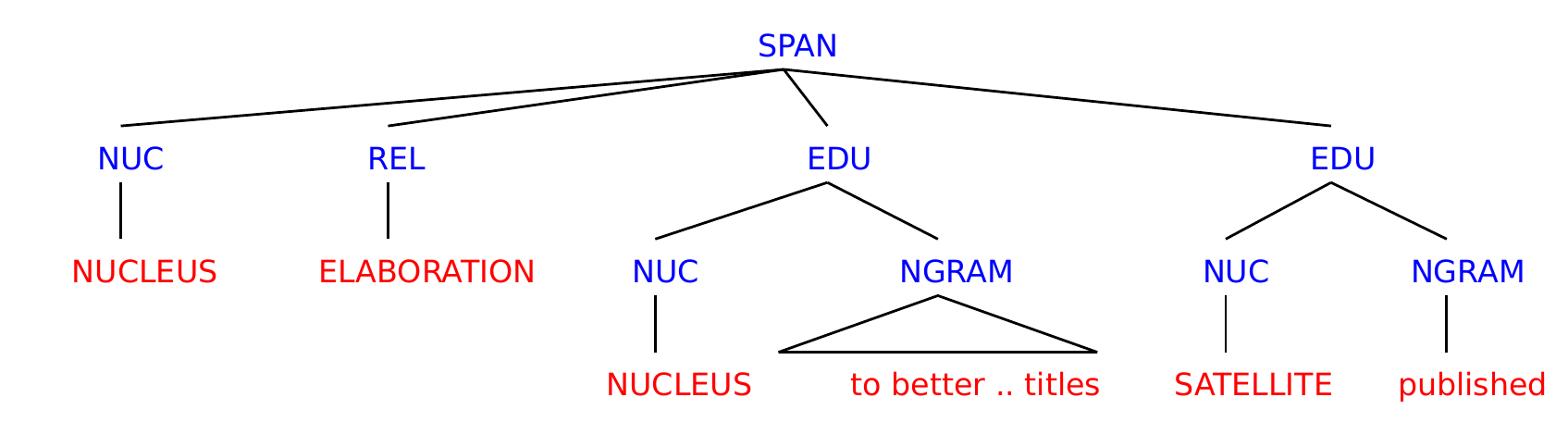}}
	    \caption{DT for \drLEXto .}
	     \label{fig:lex2}
	\end{subfigure}
	\hspace{0.3cm}
	\begin{subfigure}{.5\linewidth}
	\mbox{\hspace{-0.68cm}
		 \includegraphics [scale=0.45] {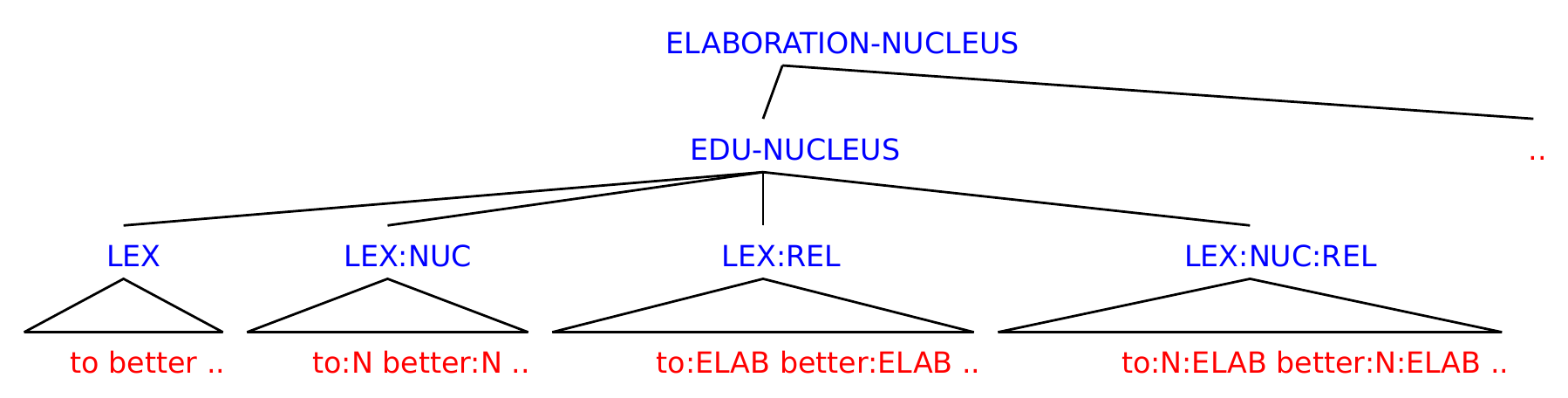}}
	    \caption{DT for \drLEXoo .}
	     \label{fig:lex1.1}
	\end{subfigure}

	\begin{subfigure}{.8\linewidth}
	\mbox{\hspace{-0.4cm}
		 \includegraphics [scale=0.48] {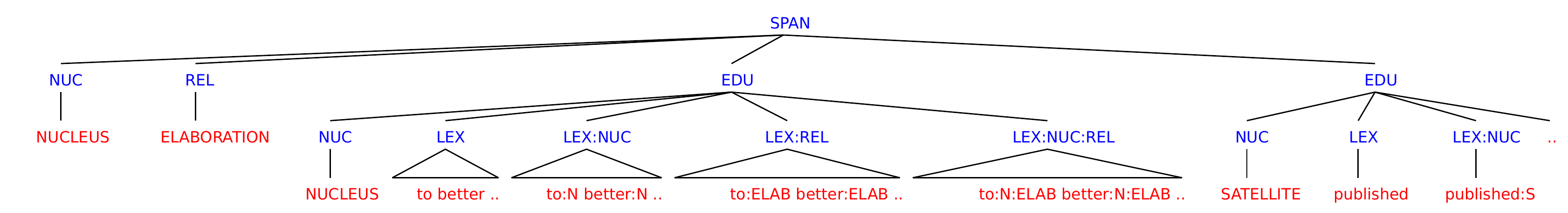}}
	    \caption{DT for \drLEXtt .}
	     \label{fig:lex2.1}
	\end{subfigure}

\caption{Five different representations of the discourse tree (DT) for the sentence \emph{ ``The new organisational structure will also allow us to enter the market with a joint offer of advertising products, to better link the creation of content for all the titles published and, last but not least, to continue to streamline significantly the business management of the company,'' added Cermak.}
Note that to avoid visual clutter, (b)--(e) show alternative representations only for the highlighted subtree in (a).
}
\label{fig:trees}
\end{figure*}

Although both \drLEX \ and \drLEXto \ allow the tree kernel to find matches at the word level, the words are compared in a bag-of-words fashion, i.e., if the trees share a common word, the kernel will find a match regardless of its position in the tree. Therefore, a word that has occurred in an EDU with status \emph{Nucleus} in one tree could be matched with the same word under a \emph{Satellite} in the other tree. In other words, the kernel based on these representations is insensitive to the nuclearity status and the relation labels under which the words are matched. \drLEXoo , an extension of \drLEX , and \drLEXtt , an extension of \drLEXto , are sensitive to these variations at the lexical level. \drLEXoo \ (Figure \ref{fig:lex1.1}) and \drLEXtt \ (Figure \ref{fig:lex2.1}) propagate the nuclearity statuses and/or the relation labels to the lexical items by including three more subtrees at the EDU level.


\section{Metric Combination and Tuning}
\label{sec:combination}
In this section, we describe our Discourse Tree Kernel (\disco) metrics.
We have two main versions:
\discolight, which combines the five DR-based metrics,
and
\discoparty, which further adds the Asiya metrics.

\subsection{\discolight}

In the previous section, we have presented several discourse tree representations that can be used to compare the output of a machine translation system to a human reference. Each representation stresses a different aspect of the discourse tree.

In order to make our estimations more robust, we propose \discolight,
a metric that takes advantage of all the previous discourse representations by linearly interpolating their scores.
Here are the processing steps needed to compute this metric:

\noindent\Ni Parsing:
We parsed each sentence in order to produce discourse trees for the human references and for the outputs of the systems.

\noindent\Nii Tree enrichment/simplification:
For each sentence-level discourse tree,
we generated the five different tree representations: \dr, \drLEX, \drLEXoo, \drLEXto, \drLEXtt.

\noindent\Niii Estimation:
We calculated the per-sentence similarity scores between tree representations of the system hypothesis and the human reference using the extended convolution tree kernel as described in the previous section. To compute the system-level similarity scores, we calculated the average sentence-level similarity;
note that this ensures that our metric is ``the same'' at the system and at the segment level.

\noindent\Niv Normalization:
In order to make the scores of the different representations comparable,
we performed a $\min$--$\max$ normalization\footnote{Where $x'=(x-\min)/(\max-\min)$.}
for each metric and for each language pair.

\noindent\Nv Combination:
Finally, for each sentence,
we computed \discolight~ as the
average of the normalized similarity scores
of the different representations.
For system-level experiments,
we performed linear interpolation of system-level scores.

\subsection{\discoparty}

One of the weaknesses of the above discourse-based metrics is that they use unigram lexical information,
which does not capture reordering.
Thus, in order to make more informed and robust estimations,
we extended \discolight \ with the composing metrics of the \asiya's ULC~metric~\cite{AsiyaMTjournal:2010},
which is a uniform linear combination of twelve individual metrics
and was the best-performing metric at the system and at the segment levels
at the WMT08 and WMT09 metrics tasks.

In order to compute the individual metrics from ULC,
we used the \asiya\ toolkit,\footnote{http://nlp.lsi.upc.edu/asiya/}
but we departed from \asiya's ULC
by replacing TER and Meteor with newer versions thereof
that take into account synonymy lookup and paraphrasing (`TERp-A' and `Meteor-pa' in \asiya's terminology).
We then combined the five components in \discolight\ and the twelve individual metrics from ULC;
we call this combination \discoparty.

We combined the scores using linear interpolation in two different ways:

\noindent\Ni \emph{Uniform combination} of $\min$-$\max$ normalized scores at the segment level.
We obtained system-level scores by computing the average over the segment scores.

\noindent\Nii \emph{Trained interpolation at the sentence level}.
We determined the interpolation weights for the above-described combination of 5+12 = 17 metrics
using a pairwise learning-to-rank framework and classification with logistic regression,
as we had done in \cite{discoMT:acl2014}.
We obtained the final test-time sentence-level scores by passing the interpolated raw scores through a sigmoid function.
In contrast, for the final system-level scores, we averaged the per-sentence interpolated raw scores.

We also tried to learn the interpolation weights at the system level,
experimenting with both regression and classification.
However, the amount of data available for this type of training was small,
and the learned weights did not perform significantly better than the uniform combination.

\subsection{Post-processing}
Discourse-based metrics, especially \dr, tend to produce many ties when there is not enough information to do complete discourse analysis. This contributes to lower $\tau$ scores for \discolight.
To alleviate this issue, we used a simple tie-breaking strategy, in which ties between segment scores for different systems are resolved by using perturbations proportional to the global system-level scores produced by the same metric,
i.e., $x'^{seg}_{sys}$= $x^{seg}_{sys}+\epsilon*\sum_s x^s_{sys}$.
Here, $\epsilon$ is automatically chosen to avoid collisions with 
scores not involved in the tie.
This post-processing is not part of the metric; it is only applied to our segment-level submission to the WMT'14 metrics task.

\section{Experimental Evaluation}
\label{sec:experiments}
In this section, we present some of our experiments to decide on the best \disco\ metric variant and tuning set. For tuning, testing and comparison, we worked with some of the datasets available from previous WMT metrics shared tasks, i.e., 2011, 2012 and 2013.
From previous experiments~\cite{discoMT:acl2014}, we know that the tuned metrics perform very well on  cross-validation for the same-year dataset. We further know that tuning can be performed by concatenating data from all the into-English language pairs, which yields better results than training separately by language pair. For the WMT14 metrics task, we investigated in more depth whether the tuned metrics generalize well to new datasets. Additionally, we tested the effect of concatenating datasets from different years. 

Table~\ref{tab:results} shows the main results of our experiments with the \disco~metrics.
We evaluated the performance of the metrics on the WMT12 and WMT13 datasets both at the segment and the system level, and we used WMT11 as an additional tuning dataset. We measured the performance of the metrics in terms of correlation with human judgements. At the segment level, we evaluated using Kendall's Tau ($\tau$),
recalculated following the WMT14 official Kendall's Tau implementation.
At the system level, we used Spearman's rank correlation ($\rho$) and Pearson's correlation coefficient ($r$). 
In all cases, we averaged the results over all into-English language pairs. The symbol `$\varnothing$' represents the untuned versions of our metrics, i.e., applying a uniform linear combination of the individual metrics.

We trained the tuned versions of the \disco\ measures using different datasets (WMT11, WMT12 and WMT13)
in order to study across-corpora generalization and the effect of training dataset size.
The symbol `+' stands for concatenation of datasets. We trained the tuned versions at the segment level using Maximum Entropy classifiers for pairwise ranking (cf. Section~\ref{sec:combination}). For the sake of comparison, the first group of rows contains the results of the best-performing metrics at the WMT12 and WMT13 metrics shared tasks and the last group of rows contains the results of the \asiya\ combination of metrics, i.e., \discoparty\ without the discourse components. 

\begin{table*}[t]
\small
\begin{center}

\begin{tabular}{r|c|c|c||cc|cc|}
\cline{3-8}
\multicolumn{2}{c|}{} & \multicolumn{2}{|c||}{Segment Level} & \multicolumn{4}{|c|}{System Level}\\
\cline{3-8}
\multicolumn{2}{c|}{} & WMT12 & WMT13 & \multicolumn{2}{|c|}{WMT12} & \multicolumn{2}{|c|}{WMT13}\\
\hline
Metric & Tuning & \multicolumn{1}{|c|}{$\tau$} & \multicolumn{1}{|c||}{$\tau$} & \multicolumn{1}{|c}{$\rho$} & \multicolumn{1}{c|}{$r$} & \multicolumn{1}{|c}{$\rho$} & \multicolumn{1}{c|}{$r$}\\
\hline
 \sempos & na & \multicolumn{1}{|c|}{--} & \multicolumn{1}{|c||}{--} & 0.902 & 0.922 & \multicolumn{1}{|c}{--} & \multicolumn{1}{c|}{--}\\
 \spede  & na & 0.254 & \multicolumn{1}{|c||}{--} & \multicolumn{1}{|c}{--} & \multicolumn{1}{c|}{--} & \multicolumn{1}{|c}{--} & \multicolumn{1}{c|}{--}\\
 {\sc Meteor-wmt13} & na & \multicolumn{1}{|c|}{--} & 0.264 & \multicolumn{1}{|c}{--} & \multicolumn{1}{c|}{--} & 0.935	& {\bf 0.950}\\
 \hline

\hline
            & $\varnothing$ & 0.171 & 0.162 & 0.884 & 0.922 & 0.880 & 0.911\\
            & WMT11         & 0.207 & 0.201 & 0.860 & 0.872 & 0.890 & 0.909\\
\discolight & WMT12         & --    & 0.200 & --    & --    & 0.889 & 0.910\\
            & WMT13         & 0.206 & --    & 0.865 & 0.871 & --    & --\\
            & WMT11+12      & --    & 0.197 & --    & --    & 0.890 & 0.910\\
            & WMT11+13      & 0.207 & --    & 0.865 & 0.871 & --    & --\\
\hline
            & $\varnothing$ & 0.257 & 0.231 & 0.907 & 0.915 & {\bf 0.941} & 0.928\\
            & WMT11         & 0.302 & 0.282 & {\bf 0.915} & {\bf 0.940} & 0.934 & 0.946\\
\discoparty & WMT12         & --    & 0.284 & --    & --    & 0.936 & 0.940\\
            & WMT13         & {\bf 0.305} & --    & 0.912 & 0.935 & -- & --\\
            & WMT11+12      & --    & {\bf 0.289} & -- & -- & 0.936 & 0.943\\
            & WMT11+13      & 0.304 & --    & 0.912 & 0.934 & -- & --\\
\hline
\hline

 & $\varnothing$            & 0.273 & 0.252 & 0.899 & 0.909 & 0.932 & 0.922\\
            & WMT11         & 0.301 & 0.279 & 0.913 & 0.935 & 0.934 & 0.944\\
{\sc \asiya}            & WMT12         & --    & 0.277 & --    & --    & 0.932 & 0.938\\
& WMT13         & 0.303 & --    & 0.908 & 0.932 & --    & --\\
            & WMT11+12      & --    & 0.277 & --    & --    & 0.934 & 0.940\\
            & WMT11+13      & 0.303 & --    & 0.908 & 0.933 & --    & --\\

\hline
\end{tabular}
\end{center}
\caption{\label{tab:results}Evaluation results on WMT12 and WMT13 datasets at segment and system level for the main combined \disco\ measures proposed in this paper.}
\vspace{-5pt}
\end{table*}

Several conclusions can be drawn from Table~\ref{tab:results}.
First, \discoparty\ is better than \discolight\ in all settings, indicating that the discourse-based metrics are very well complemented by the heterogeneous metric set from \asiya. \discolight\ achieves competitive scores at the system level (which would put the metric among the best participants in WMT12 and WMT13); however, as expected, it is not robust enough at the segment level. On the other hand, the tuned versions of \discoparty\ are very competitive and improve over the already strong \asiya\ in each configuration both at the segment- and the system-level. The improvements are small but consistent, showing that using discourse increases the correlation with human judgments.

Focusing on the results at the segment level, it is clear that the tuned versions offer an advantage over the simple uniform linear combinations. Interestingly, for the tuned variants, given a test set, the results are consistent across tuning sets, ruling out over-fitting; this shows that the generalization is very good. This result aligns well with what we observed in our previous studies~\cite{discoMT:acl2014}. Learning with more data (WMT11+12 or WMT12+13) does not seem to help much, but it does not hurt performance either. Overall, the $\tau$ correlation results obtained with the tuned \discoparty\ metric are much better than the best results of any participant metrics at WMT12 and WMT13 (20.1\% and 9.5\% relative improvement, respectively).

At the system level, we observe that tuning over the \discolight\ metric is not helpful (results are actually slightly lower), while tuning the more complex \discoparty\ metric yields slightly better results. 

The scores of our best metric are higher than those of the best participants in WMT12 and WMT13, according to Spearman's $\rho$, which was the official metric in those years.
Overall, our metrics are comparable to the state-of-the-art at the system level. The differences between Spearman's $\rho$ and Pearson's $r$ coefficients are not dramatic, with $r$ values being always higher than $\rho$.

Given the above results, we submitted the following runs to the WMT14 Metrics shared task: \Ni \discoparty\ tuned on the concatenation of datasets WMT11+12+13, as our primary run; \Nii Untuned \discoparty, to verify that we are not over-fitting the training set; and \Niii Untuned \discolight, to see
the performance of a metric using discourse structures and word unigrams.

The results for the WMT14 Metrics shared task have shown that our primary run, \discoparty\ tuned,
was the \emph{best-performing} metric both at the segment- and at the system-level~\cite{machacek-bojar:2014:WMT}.
This metric yielded significantly better results than its untuned counterpart,
confirming the importance of weight tuning and the absence of over-fitting during tuning.
Finally, the untuned \discolight \ achieved relatively competitive, albeit 
slightly worse
results for all language pairs, except for Hindi-English,
where system translations resembled a ``word salad'', and were very hard to discourse-parse accurately. 

\section{Conclusion}
\label{sec:conclusion}
We have presented experiments with novel automatic metrics for machine translation 
evaluation that take discourse structure into account.
In particular, we used RST-style discourse parse trees, which we compared using convolution kernels.
We further combined these kernels with metrics from \asiya, also tuning the weights.
The resulting \discoparty \ tuned metric was the best-performing at the segment- and system-level 
at the WMT14 metrics task.

In an internal evaluation on the WMT12 and WMT13 metrics datasets, this tuned combination 
showed correlation with human judgments that outperforms the best systems that participated 
in these shared tasks. The discourse-only metric 
ranked near the top at the system-level 
for WMT12 and WMT13; however, it is weak at the segment-level since it is sensitive to parsing errors,
and 
most sentences have very little internal discourse structure.

In the future, we plan to work on an integrated representation of syntactic, semantic and
discourse-based tree structures, which would allow us to design evaluation
metrics based on more fine-grained features, and would also allow us to train such metrics using kernel methods. Furthermore, we want to make use of discourse parse information beyond the sentence level.



\bibliographystyle{acl}
\bibliography{biblio-reduced}

\end{document}